\documentclass[10pt,twocolumn,letterpaper]{article}
\usepackage{iccv}
\usepackage[accsupp]{axessibility} 
\usepackage{times}
\usepackage{epsfig}
\usepackage{graphicx}
\usepackage{amsmath}
\usepackage{amssymb}
\usepackage{cuted}
\usepackage{capt-of}
\usepackage{authblk}
\usepackage{caption}
\usepackage[normalem]{ulem}
\usepackage{cancel}
\usepackage[belowskip=-14pt,aboveskip=0pt]{caption}
\newcommand{\cmmnt}[1]{}

\usepackage[breaklinks=true,bookmarks=false]{hyperref}

\iccvfinalcopy 

\def\iccvPaperID{7102} 

\ificcvfinal\fi

\begin{document}

\title{STRIVE: Scene Text Replacement In Videos \vspace{-1em}}

\author[1]{Vijay Kumar B G }
\author[2]{Jeyasri Subramanian}
\author[3]{Varnith Chordia}
\author[2]{Eugene Bart}
\author[4]{Shaobo Fang}
\author[5]{\\Kelly Guan}
\author[3]{Raja Bala}
\affil[1]{NEC Laboratories, America}
\affil[2]{Palo Alto Research Center \thanks{email ids: bg.vijay.k@gmail.com, jsubrama@parc.com, vchordia@amazon.com, bart@parc.com, sfang.ee@gmail.com, rajabl@amazon.com}}
\affil[3]{Amazon}
\affil[4]{Work done at PARC}
\affil[5]{Stanford University}

\ificcvfinal\thispagestyle{empty}\fi

\maketitle
\vspace*{-1cm}
\begin{abstract}
    We propose replacing scene text in videos using deep style transfer and learned photometric transformations. Building on recent progress on still image text replacement, we present extensions that alter text while preserving the appearance and motion characteristics of the original video. Compared to the problem of still image text replacement, our method addresses additional challenges introduced by video, namely effects induced by changing lighting, motion blur, diverse variations in camera-object pose over time, and preservation of temporal consistency. We parse the problem into three steps. First, the text in all frames is normalized to a frontal pose using a spatio-temporal transformer network. Second, 
    the text is replaced in a single reference frame using a state-of-art still-image text replacement method. Finally, the new text is transferred from the reference to remaining frames using a novel learned image transformation network that captures lighting and blur effects in a temporally consistent manner. Results on synthetic and challenging real videos show realistic text transfer, competitive quantitative and qualitative performance, and superior inference speed relative to alternatives. We introduce new synthetic and real-world datasets with paired text objects. To the best of our knowledge this is the first attempt at deep video text replacement.  

\end{abstract}

\begin{figure}[t]
\centering
\includegraphics[width=6.5cm,height=4cm]{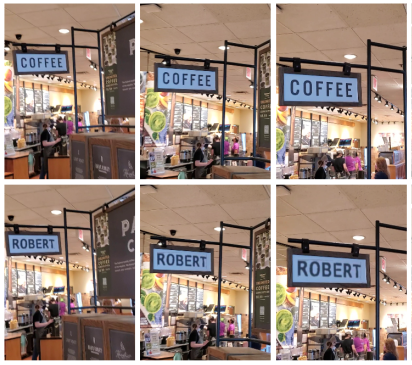} 
\caption{Our method replaces the scene text in the original video ("COFFEE" in upper row) with a personalized string ("ROBERT" in bottom row) while preserving the original geometry, appearance, and temporal properties. 
}
\label{fig:cover_fig}
\end{figure}

\section{Introduction}
\begin{figure*}[t]
\centering
\includegraphics[width=18cm]{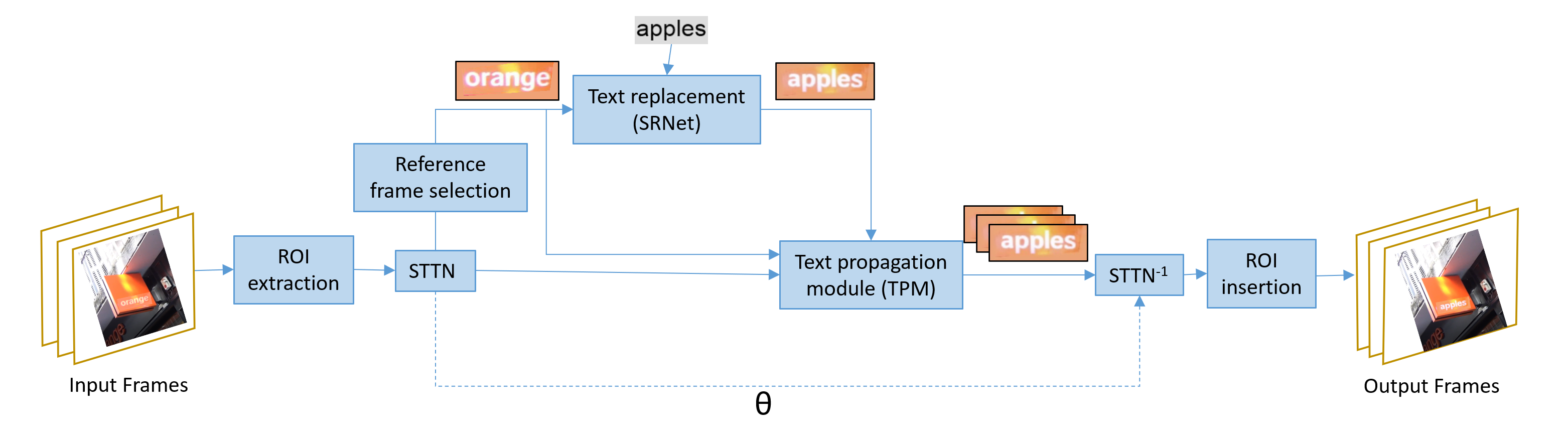}
\caption{Overview of STRIVE. Text region of interest (ROI) is extracted from each frame and frontalized using a spatiotemporal transformation network (STTN) with parameters $\theta$. Next a reference frame is selected and text is replaced using SRNet. The new text is transferred to other frames via a novel text propagation module (TPM), and reinserted into the frame after reverting to the original pose via $STTN^{-1}$.}
\label{fig:overviewFig}
\end{figure*}

We address the problem of realistically altering scene text in videos. Our primary application is to create personalized content for marketing and promotional purposes. An example would be to replace a word on a store sign with a personalized name or message, as shown in Figure  \ref{fig:cover_fig}. Other applications include language translation, text redaction for privacy, and augmented reality. For research purposes, the ability to realistically manipulate scene text also enables augmentation of datasets for training text detection, recognition, tracking, erasure, and adversarial attack detection. Traditionally text editing in images is performed manually by graphic artists, a process that typically entails a long painstaking process to ensure the original geometry, style and appearance are preserved. For videos, this effort would be considerably more arduous. \\
Recently several attempts have been made to automate text replacement in still images based on principles of deep style transfer (\cite{stefann2020}, \cite{mcgan2018}, \cite{tetgan2019}, \cite{srnet2019}, \cite{swaptext2020}). We leverage this progress to tackle the problem of text replacement in videos. In addition to the challenges faced in the still image case, video text replacement must respect temporal consistency and model effects such as lighting changes, blur induced by camera and object motion. Furthermore, over the course of the video, the pose of the camera relative to the text object can vary widely, and hence text replacement must be able to handle diverse geometries. A logical method to solving the problem would be to train an image-based text style transfer module on individual frames, while incorporating temporal consistency constraints in the network loss. The problem with such an approach is that the network performing text style transfer (an already non-trivial task) is now additionally burdened with handling  geometric and motion-induced effects encountered in video. We show in experiments that current still-image text replacement techniques such as \cite{srnet2019, pix2pix2017} do not robustly handle such effects. We therefore take a different approach. We first extract text regions of interest (ROI) and train a spatio-temporal transformer network (STTN) to frontalize the ROIs in a temporally consistent manner. We then scan the video and select a reference frame with high text quality, measured in terms of text sharpness, size, and geometry. We perform still-image text replacement on this frame using a state-of-art method SRNet \cite{srnet2019} trained on video frames. We then transfer the new text onto other frames with a novel \textit{text propagation module} (TPM) that takes into account changes in lighting and blur effects with respect to the reference frame. TPM takes as input the reference and current frame from the original video, infers an image transformation between the pair, and applies it to the altered reference frame generated by SRNet. Crucially, TPM takes temporal consistency into account when learning pairwise image transforms. 
Our framework, dubbed STRIVE (\textbf{S}cene \textbf{T}ext \textbf{R}eplacement \textbf{I}n \textbf{V}id\textbf{E}os) is summarized in Figure  \ref{fig:overviewFig}.\\
To our knowledge this is the first attempt at replacing scene text in videos. We make the following contributions:\\
1) A modular pipeline that disentangles the problem of text replacement in a single reference frame from modeling the flow of the replaced text within the scene over time. Such a parsing of the problem into simpler subtasks serves to both simplify training and reduce computations during inference. \\
2) A learned parametric differential image  transformation that captures temporal photometric changes between a pair of aligned ROIs in the original video, and applies it to ROIs in the text-altered video. The transform consists of a learnable blur/sharpness operator, and is trained on synthetic data and fine-tuned via self-supervision on real-world images. \\
3) New synthetic and real-world datasets comprising a diversity of annotated scene text objects in videos. A subset of the videos comprise triplets of ROIs with aligned source text, target text, and plain background. The datasets are available at \url{https://striveiccv2021.github.io/STRIVE-ICCV2021/}. 
\section{Related Work}
\subsection{Style Transfer}
Our approach has its roots in deep neural style transfer \cite{gatys2016}, \cite{huang2017arbitrary}. Specifically pix2pix \cite{pix2pix2017} forms the backbone for image-to-image transfer tasks and will be used as a baseline in our experiments. Extensions to style transfer for videos have been proposed by several researchers, whereby temporal consistency constraints are added to the style losses applied on individual frames \cite{huang2017}, \cite{gupta2017}. We disentangle style transfer from geometric and photometric variations encountered in video, and apply temporal consistency constraints on the latter.

\subsection{Scene Text Detection, Recognition}
Text replacement relies upon successful text detection and recognition. As expected, deep learning methods define the state of art for both image and video input \cite{east2017},\cite{wang2019},\cite{cheng2019}. We use the video text detection module within the Amazon \textit{ReKognition} toolbox to extract text ROIs for our pipeline. Related efforts to synthesize realistic scene text for training models include \cite{zhan2019}, \cite{long2020}, \cite{Zhan_2019_CVPR}.

\subsection{Scene Text Replacement}
Recently, deep neural techniques have been proposed for editing scene text in still images. These fall into two categories: techniques that replace individual characters (\cite{stefann2020}, \cite{mcgan2018}, \cite{tetgan2019}, \cite{yang2017}, \cite{yang2019}), and those that replace entire words (\cite{srnet2019}, \cite{swaptext2020}). While the approaches vary in the types of effects that can be modeled, the main steps are: i) inpainting to erase existing text \cite{tursun2019}; ii) transfer of input text style to new characters or words; iii) a fusion step to combine foreground and background regions for realistic outputs. 
Since it is difficult to acquire paired text data in real world scenes, existing methods are trained on synthetic datasets. The closest application we are aware of is the camera mode in the Google Translate app, which overlays language translations on scene text in the camera preview mode. Since the primary purpose is translation, there is no attempt to strictly match the appearance of the original text. 

\subsection{Learning Image Transformations}
Our text propagation module (TPM) learns photometric transformations between a reference and a non-reference frame. Several works have tackled a similar problem in the context of image enhancement. Gharbi et al. \cite{gharbi2017} learn to automatically enhance images by training a deep bilateral transform network from paired  (pre- and post-enhancement) training data. Related works \cite{gharbi2015}, \cite{chen2016} learn to predict a parameterized transformation from a pair of low-resolution images and apply this to the high resolution version. The transform models complex photo enhancement operators with a "recipe" of local affine transforms. We draw inspiration from these techniques to learn a parametric transform between reference and non-reference frames in the original video, with several key differences. First, in previous works, the transformation learned from image $I_1$ to its enhanced version $I_2$ is then applied on a high-resolution version of $I_1$. Each transform is thus intimately tied to a specific image. In our case the transform learned from an image pair with source scene text is applied to a \emph{different} image with target text, and hence its structure must be abstracted from the original image content. Secondly, while previous efforts are aimed at learning image enhancement operations, our purpose is to model changes in lighting and camera motion encountered across video frames, which often include distortions such as image blurring. Finally our transformations must exhibit temporal consistency, which is not applicable for still-image transforms.

\subsection{Image blur estimation and correction}
Techniques have been proposed for using deep CNNs to estimate blur kernels from images and videos with the goal of blind image deblurring \cite{yan2016}, \cite{wu2019}, \cite{zhang2020}. Our blur estimation in TPM adopts a similar approach, however with a different purpose of estimating a \emph{differential} blur transform between a pair of images, rather than estimating blur in an absolute sense from a single image. 

\section{Methodology}

Referring to Fig. \ref{fig:overviewFig}, we first extract tight ROIs (i.e. bounding boxes) for the source text from the input video using the Amazon \textit{Rekognition} API. All operations described next are carried out on the ROIs.  
\subsection{Reference Frame Selection} \label{subsubsec:RFS}
We select a single reference frame for text replacement. To ensure successful text style transfer, we desire a frame wherein the source text clearly legible, of high contrast, and maximally frontal in pose. 
We compute four metrics on the text ROI. 1) Only samples with optical character recognition (OCR) confidence greater than 0.99 (as reported by \textit{ReKognition}) are considered. This eliminates excessively blurry, distorted or occluded text objects.  2) Image sharpness is measured as the variance of the Laplacian of the luminance image \cite{pertuz2013}, and the top 10 frames with the highest sharpness scores are selected. 3) The image is binarized using Otsu's algorithm \cite{otsu1979} and the normalized interclass variance $s_1$ between foreground and background regions is obtained as a measure of text contrast. 4) The ratio $s_2$ of the area of the tight ROI bounding box to the area of the subsuming axis-aligned rectangle is calculated as an estimate of frontal pose.  A composite text quality score is computed as $\alpha_1s_1 + \alpha_2s_2$ and the frame with the highest score is selected as the reference frame. Both $\alpha_1$ and $\alpha_2$ are heuristically chosen based on visual evaluation of replaced text on random videos.

\subsection{Pose Normalization} 
Text objects in videos can undergo diverse geometric distortions due to varying object pose relative to the camera. To minimize the effect of this distortion on text style transfer, we normalize and align the source text in all ROIs to a canonical frontal pose before the replacement operation. This enables both robust text style transfer on the reference frame, and the propagation of the replaced text onto the remaining frames via learned image transforms. We make a simplifying assumption that the scene text is on a planar surface. This covers many common cases such as street and store signage, banners, etc. Under the planar assumption, pose alignment is accomplished via a perspective transform. We adopt and extend the Spatial Transformer Network \cite{stn2015} (STN) as a learned approach to perspective correction that is computationally efficient at runtime. STN predicts the parameters $\theta$ of a geometric correction transform via a localization network, and applies the transform to the image via grid generation and resampling operators. The original STN is trained as part of a supervised classification task. We adopt the same network architecture, but instead explicitly train it to produce temporally consistent frontal ROIs. We obtain binary masks of text ROIs with Mask R-CNN \cite{He_2017_ICCV}. The training samples comprise a stack of distorted input ROI masks and a frontal ROI mask of synthetic text serving as the target label.  The net is trained on the following loss:
\begin{equation}
\label{eq:STTN_Loss}
	L_{STTN}=L_{\theta}+\lambda_1L_{pix}+\lambda_2L_{t}
\end{equation}
where the first two terms are from the original model: $L_{\theta}$  is the mean-squared error (MSE) between the true and predicted homography parameter vector $\theta$, and $L_{pix}$ is the pixelwise MSE between predicted and true text ROIs.  We use $L_2$ norm for the above losses following the practice of Nguyen \etal  \cite{Nguyen18} and DeTone \etal \cite{Detone16}; this choice produced good results in our experiments. Additionally we introduce temporal consistency loss $L_t$, defined as the MSE between $\theta$ for adjacent video frames: $L_{t}=\sum_{j} \lvert \theta_i - \theta_j \rvert ^2$, where index $i$ denotes current frame and the summation is over a number of neighboring frames $j \neq i$. This term ensures that perspective correction parameters vary smoothly over adjacent frames.  Fig. \ref{fig:overviewFig} shows how the spatiotemporal transform network ($STTN$) is incorporated into the overall framework. After text replacement, the ROI is sent through the inverse perspective transform $(STTN^{-1})$ to restore it to the original scene geometry and inserted into the original frames to produce the output video.

\subsection{Text Replacement in Reference Frame}
We select SRNet \cite{srnet2019} to replace text in the reference frame. In principle, any state-of-art still-image text replacement technique can be used. SRNet takes the input ROI and the target text in the form of a mask, and executes sub-networks for background generation, foreground creation via transfer of style from source to target text, and blending of background and foreground to produce the target text ROI. We train SRNet on video frames from our datasets. During training we introduce additional augmentations for perspective distortion, and motion and out-of-focus blur encountered in videos.

\subsection{Text Propagation}
\begin{figure}[t]
\centering
\includegraphics[width=8.5cm]{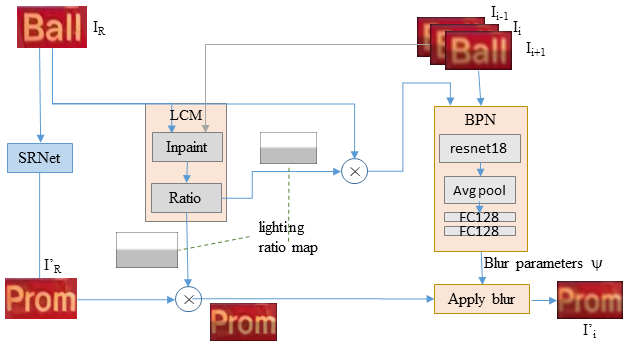}
\caption{Text Propagation Module (TPM) derives a local image transform between reference and non-reference ROIs in the original video ("Ball") and applies the transform to the reference ROI ("Prom") in the altered video. The transform comprises a Lighting Correction Module (LCM) followed by a Blur Prediction Network (BPN).}
\label{fig:TPNet}
\end{figure}
The main novel element of our work is text propagation from reference frame to the rest of the video via TPM. Our key insight is to avoid repeatedly performing text replacement on every frame, and instead to solve the simpler problem of learning the changes in text appearance over the video. 
We posit that the image transformation between two video frames within a localized text ROI can be adequately modeled in terms of simple parametric changes in lighting and image sharpness due to changes in camera/object/lighting conditions over time. An advantage of our approach is that we are able to use self-supervision to learn the parameters of the model without relying on a large set of paired videos and labels. In detail, let $I_R$ and $I_i$ be respectively the ROIs from the reference and $i^{th}$ frame containing the source text in the input video. Similarly let $I'_R$ and $I'_i$ be corresponding ROIs of the target text in the output video. All ROI's are pose-corrected and aligned by STTN, and are scaled to a fixed canonical image dimension prior to TPM processing. A parametric transform is learned between $I_R$ and $I_i$, and is then applied to $I'_R$ to predict $I'_i$. This transform comprises two stages, a lighting correction module (LCM) and differential blur prediction network (BPN), as shown in Fig. \ref{fig:TPNet}.\\
The LCM captures appearance differences between reference and current ROI due to changes in illumination, including shadows and shading. Since the color of an object is a product of its reflectance and the illumination, we surmise that to first order, changes in light reflected off a fixed text object can be modeled by independent channel-wise scaling of R, G, and B channels in a spatially varying manner. Namely, changes in lighting between two aligned ROIs can be obtained from their ratio $I_i/I_R$, which is then multiplied by $I'_R$ to obtain the lighting-corrected output for $I'_i$. In practice despite the fact that $I_R$ and $I_i$ are aligned via STTN, even minor imperfections in alignment can result in gross errors in the ratio map particularly around text edges. Such errors become even more noticeable when applying ratio correction to \emph{new} text in $I'_R$. To address this issue, we assume that scene text is commonly placed on a smooth background, and apply inpainting to obtain estimates of the plain background, denoted $I_{Rp}$ and $I_{ip}$. We use the deep inpainting module within SRNet for this purpose. The ratio of the inpainted versions defines a multiplicative correction to $I'_R$ (see Fig. \ref{fig:TPNet}). In practice we compute the ratio $(I_{ip}+\epsilon)/(I_{Rp}+\epsilon)$, where a small $\epsilon$ avoids singularities near zero. The implicit assumption with the ratio model is that the reflective properties of the foreground text and background are similar. Further, to ensure temporal robustness we compute a weighted average of inpainted ROIs over $N$ neighboring frames before computing the ratio. The latter is multiplied by both the original and altered reference frames $I_R$ and $I'_R$ to produce lighting-corrected versions that are then passed to the blur prediction network.\\\\
BPN is a novel CNN-based method for predicting a transformation between a pair of images that can result from spatial aberrations, including motion blur, out-of-focus blur, and resolution differences due to varying distances between camera and text object. We model possible frame-to-frame distortions within a local text ROI using the following transformation:
\begin{equation}
\label{eq:blurmodel}
	I_{i}(x,y)=(1+w)I_R(x,y) - wI_R(x,y)*G_{\sigma,\rho}(x,y)
\end{equation}
where $w \in [-1, 1]$ and  $G_{\sigma,\rho}$ is an oriented 2D Gaussian filter rotated by angle $\rho$:
\begin{equation}
\label{eq:gaussian}
	G_{\sigma,\rho}(x,y) = Ke^{-(\frac{x'^2}{\sigma_{x}^{2}} + \frac{y'^2}{\sigma_{y}^{2}})}
\end{equation}
where $K$ is a normalizing constant, $x' = x\cos{\rho} + y\sin{\rho}$ and $y' = -x\sin{\rho} + y\cos{\rho}$. The case of $\sigma_x \approx \sigma_y$ yields an isotropic point spread function (PSF) modeling out-of-focus blur and resolution differences, while a significant difference between these two parameters models an anisotropic blur in direction $\rho$, encountered with typical camera or object motion. As $w$ varies from -1 to 0 to 1, the transformation varies from image sharpening to identity to image blurring. (Image sharpening with $w < 0$ is occasionally required if the current frame is sharper than the reference frame.) BPN takes one reference frame $I_R$ and a sliding window of $N$ frames ${I_i}$ around the current ($i$-th) time instance, and predicts $N$ sets of parameters $\psi = [\sigma_x, \sigma_y, \rho, w]$ that best transforms $I_R$ to the $N$ output frames (with respect to spatial frequency characteristics) via the blur model of Eqn. (\ref{eq:blurmodel}). The network thus takes in $N$+1 image inputs and predicts 4$N$ parameters. Predicting transforms on frame stacks ensures temporal consistency. The network architecture comprises a ResNet18 backbone \cite{resnet2016}, an average pooling layer and two fully connected layers, as shown in Fig. \ref{fig:TPNet}. The training loss is defined as:
\begin{equation}
\label{eq:bpnloss}
	L_{BPN}=\lambda_{\psi}L_{\psi}+\lambda_{R}L_{R}+\lambda_TL_{T}
\end{equation}
where $L_\psi$ is the squared error loss between the true and predicted parameter vectors $\psi$; $L_R$ is the  mean-squared image reconstruction error between predicted and true ROIs $I_i$; and $L_T$ is a temporal consistency loss that discourages large fluctuations in $\psi$ over time: $L_{T}=\sum_{j} \lvert \psi_i - \psi_j \rvert ^2$, where index $i$ denotes current frame and the summation is over $N$-1 neighboring frames $j \neq i$. The predicted $I_i$ is obtained by applying the blur model in Eqn (\ref{eq:blurmodel}) and (\ref{eq:gaussian}) with parameters $\psi$ to $I_R$. Note that the blur model is differentiable w.r.t. $\psi$ and thus can be applied within the training loop.\\
\begin{figure}[t]
\centering
\includegraphics[width=8.5cm,height=5cm]{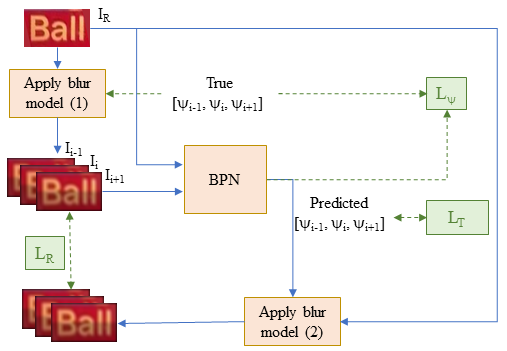}
\caption{Training of BPN. In Stage 1, a blur model (1) with known parameters is applied to reference $I_R$. The input and outputs are passed to BPN which learns to regress the parameters on losses $L_{\psi}$, $L_R$, $L_T$. In Stage 2, frame tuples from real videos are passed to BPN which regresses parameters via self-supervision on $L_R$ and $L_T$ applying blur model (2). $N$=3 in this illustration.  }
\label{fig:BPN_training}
\end{figure}
BPN is trained in two stages, as illustrated in Fig. \ref{fig:BPN_training}. In the first stage, Eqn (\ref{eq:blurmodel}) with known parameters $\psi$ is applied to reference ROIs $I_R$ from  synthetic videos  to obtain training pairs ($I_R, I_i$). In this phase, all three loss terms in Eqn (\ref{eq:bpnloss}) are minimized. As part of augmentation, inputs $I_i$ are warped with translational jitter in $x$ and $y$ directions to immunize the network to minor mis-alignments between $I_i$ and $I_R$ encountered in real video frames. In the second phase, BPN is fine-tuned via self-supervision with ($I_R, I_i$) tuples extracted from real-world videos. Here only $L_R$ and $L_T$ are minimized since the true $\psi$ are unknown. During inference, the ROI pair ($I_R, I_i$) from the original video is passed through BPN, and the predicted parameters are used to apply Eqn (\ref{eq:blurmodel}) to the altered ROI $I'_R$ to produce the ROI $I'_R$ for the current ($i$-th) frame, as shown in Fig. \ref{fig:TPNet}.

\section{Experimental Results}
To our knowledge there are no existing datasets or benchmarks for the problem of video text replacement. We thus evaluate our technique against still image replacement baselines trained and applied on video frames. Note that advances in still-image replacement are an enabler rather than competitor to STRIVE, since we rely on the still transfer for the reference frame.\\ 

\subsection{Datasets and Experimental Setup}
\begin{figure}[t]
\centering
\includegraphics[width=8cm]{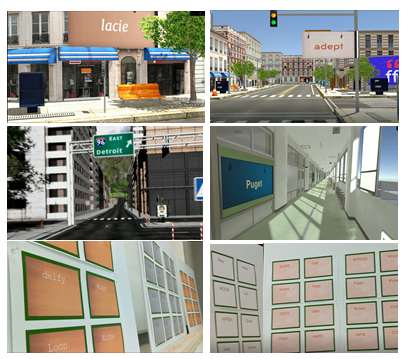}
   \caption{Example video frames from $Synthtext$ (top 2 rows) and $Robotext$ (bottom row).}
\label{fig:datasets}
\end{figure}

1. $Synthtext$: We have developed a dataset of 120 synthetic videos using the \textit{Unity} simulator. Indoor and outdoor scenes are modeled with diverse text styles against varied backgrounds, captured with different lighting, camera and object motion characteristics. A given scene is generated with multiple text strings, providing many source-target pairs for training and testing our models. Unlike existing synthetic datasets comprising clean frontal ROIs, our video simulations produce ROIs with realistic geometric and photometric distortions, including motion blur and shadows. Fig. \ref{fig:datasets} shows examples.

2. $Robotext$: We acquire first-of-a-kind videos captured by a Google Pixel 2 XL smartphone mounted on a $Create 2$ robotic platform. The robot is programmed to travel in random trajectories around a gallery of text posters mounted in a large indoor hall. Trajectories include linear and curved paths at varying travel speeds. The posters are designed such that different words of the same style and background are adjacent to each other, to train text replacement models, as shown in Fig. \ref{fig:datasets}. This dataset comprises approximately 5000 short video clips.\\
3. $Realworld$: We have collected challenging real-world videos in indoor and outdoor environments including camera motion related to walking and driving. We drew from two sources. The first is the "Text in Videos" dataset from the ICDAR 2015 Robust Reading Competition \cite{icdar2015} comprising 25 videos of scene text captured in the wild.  
From these, we curated a subset of 15 videos of sufficient quality for our task. Additionally we collected our own dataset of 22 videos of similar scene content and diversity as the ICDAR set.\\
All datasets are annotated with bounding boxes around text objects in each frame using AWS \textit{ReKognition} software.\\
Pix2pix and SRNet are trained on frames from 1000 video clips from $Synthtext$ and $Robotext$ following the protocols of the original implementations \cite{pix2pix2017}, \cite{srnet2019}, with additional augmentations for pose variations.  STTN is trained on frames from 100 video clips from $Synthtext$ and $Robotext$ with frontal stylized text masks serving as training targets. For reference frame selection, parameters $\alpha_1$ and $\alpha_2$ (see Section \ref{subsubsec:RFS}) were selected as 0.7 and 0.3 based on cross-validation experiments. Stage 1 of BPN is trained on 100 videos from $Synthtext$, with 10\% of the training data used for validation. Stage 2 is trained on 900 videos from $Robotext$ and a subset of 25 videos from $Synthtext$. Both stages are trained with 100 epochs. We use a neighborhood of $N$=3 consecutive frames for imposing temporal consistency. The ADAM optimizer is used with a learning rate of 0.0005 for stage 1 and 0.0003 for stage 2. All implementations are in the Pytorch framework \cite{Pytorch15} with GPU acceleration, and all tests reported below are performed on independent datasets. 

\subsection{Evaluation Metrics}
\noindent 1. We compute MSE, peak signal-to-noise ratio (PSNR), and average structural similarity (SSIM) scores \cite{ssim2004} between estimated and ground-truth ROIs for frames from synthetic test videos. \\
2. We evaluate OCR accuracy of \textit{ReKognition} on video text in real scenes. We measure the number of word-level errors on the target text as a fraction of the number of frames in which the source text was correctly recognized.\\
3. We evaluate high-frequency jitter of the target text by analyzing bounding box coordinates returned by \textit{ReKognition} from altered videos. Let a box vertex have time-varying coordinates [$x(t)$,$y(t)$]. We extract a high pass signal [$\Tilde{x}(t)$,$\Tilde{y}(t)$] by subtracting a lowpass-filtered version, and computing the root mean of $\Tilde{x}^2(t)+\Tilde{y}^2(t)$  as a measure of temporal jitter.\\
4. We perform timing analysis, simulating the personalized marketing scenario where a single input video is used to generate $K$ altered output copies, each with a different target text string. We measure average frame rate over 20 randomly selected input videos on a Linux machine running Ubuntu 16.04 with a single GeForce GTX GPU.     
\subsection{Quantitative Analysis}
\begin{figure}[!t]
\centering
\includegraphics[width=4cm]{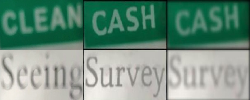}
   \caption{BPN ablation: input ROI (left), output ROI without BPN (middle), output with BPN (right).}
\label{fig:bpn_ablation}
\end{figure}

Reconstruction performance on synthetic videos is shown in the first 3 columns of Table \ref{tab:quant_mse}. Our first result is an ablation study on the effect of reference frame selection (disabled by always selecting the first frame as the reference). We see that this step has a noticeable effect in reconstruction accuracy (``w/o ref frame" in Table \ref{tab:quant_mse}). As seen in Fig. \ref{fig:comparisons_synth}, SRNet struggles to correctly replace small or strongly distorted text, and our reference frame selection criteria avoid selecting such frames for text replacement. Next we examine the impact of BPN (disabled by setting $w=0$ in Eq. (\ref{eq:blurmodel}). As seen in  Table \ref{tab:quant_mse}, BPN also plays a crucial role (``w/o BPN" in Table \ref{tab:quant_mse}). The most pronounced benefit is seen for videos that contain significant motion or focus blur, as shown in Fig. \ref{fig:bpn_ablation} where BPN accurately models focus blur. 

Next, we compare STRIVE with two competing alternatives. The first is a pix2pix baseline \cite{pix2pix2017} trained on individual video frames. To give pix2pix a fair chance at transferring text on videos, during training we supply the network with 3 consecutive frames and incorporate a temporal coherence constraint that penalizes MSE loss between the current output and its two neighboring frames. The second (and stronger) baseline is SRNet trained and applied on individual video frames. As seen in Table \ref{tab:quant_mse}, STRIVE outperforms both methods in terms of MSE, PSNR, SSIM. \\
\begin{table}
	\centering
	\setlength{\tabcolsep}{3.2pt}
	\begin{tabular}{|l|c|c|c|c|c|}
	\hline
		 Method & MSE $\downarrow$ & PSNR $\uparrow$ & SSIM $\uparrow$ & OCR $\uparrow$ & Jitter $\downarrow$ \\ \hline
		 Pix2pix \cite{pix2pix2017} & 0.0593 & 12.27 & 0.531 & - & -  \\
		 \hline
		 SRNet \cite{srnet2019} & 0.0227 & 16.44 & 0.598 & 0.771 & 5.10  \\
		 \hline
		 w/o ref frame & 0.0203 & 16.93 & 0.596 & - & -  \\
		 \hline
		 w/o BPN & 0.0203 & 16.93 & 0.594 & - & -  \\
		 \hline
		 STRIVE & \textbf{0.0186} & \textbf{17.31} & \textbf{0.604} & \textbf{0.957} & \textbf{1.48}  \\
		 \hline
	\end{tabular}\\[2ex]
\caption{Quantitative results on $Synthtext$ (first 3 columns) and $Robotext$ and $Realworld$ videos (last 2 columns). For MSE and jitter, smaller is better; for PSNR, SSIM, OCR greater is better.} 
\label{tab:quant_mse}
\end{table}
We compare STRIVE with frame-wise SRNet in terms of OCR accuracy on real and synthetic videos. From column 4 of Table \ref{tab:quant_mse}, STRIVE significantly outperforms SRNet. We owe this largely to the failure of SRNet to replace text under strong geometric distortions and photometric blur. Finally we compare temporal smoothness of the replaced text in column 5 of Table \ref{tab:quant_mse} and observe that STRIVE produces substantially less jitter, owed to temporal consistency constraints that are absent in per-frame SRNet. The jitter is clearly evident when viewing the videos (see the supplemental section).\\
In comparing the three methods, pix2pix attempts holistic transfer, accounting for text style, color, background, geometry, and lighting. SRNet decouples the transfer of foreground text from background, which helps performance. STRIVE benefits from an additional level of disentanglement, namely separation of single-image transfer from modeling of geometric and appearance variations over time. Interestingly, frame-wise SRNet offers a form of ablation, as it is essentially STRIVE without the pose normalization and text propagation step.\\
We perform inference timing tests for different output run lengths ($K$=1, 50, 100) of output videos created from a given input video. The average SRNet inference speed is 1.67 frames per second (fps) for all $K$, while STRIVE achieves faster rates of 2.11, 7.83, and 8.01 fps for $K$ = 1, 50, 100 respectively. This is because STRIVE performs expensive text replacement only on one reference frame. Furthermore, parameters for all text propagation functions (STTN, lighting, blur model in Eqn \ref{eq:blurmodel}) are computed once offline per input video, and reused in runtime for all $K$ output copies. In contrast, framewise SRNet must replace text in every frame in every video, and hence suffers from the same low throughput for all run lengths.

\subsection{BPN Simulation}
\begin{figure}[!t]
\centering
\includegraphics[width=6.5cm,height=3cm]{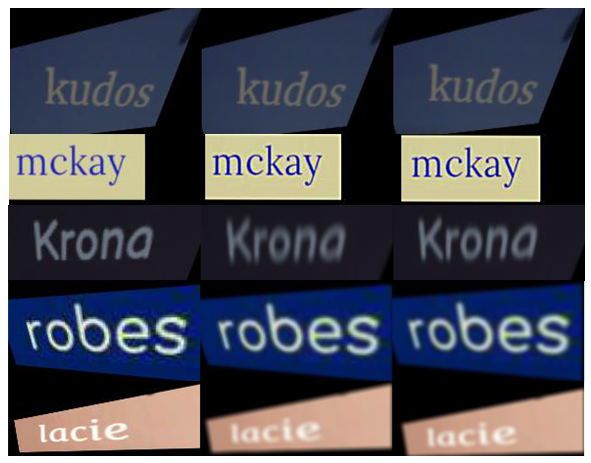}
   \caption{Results of BPN simulation. From top to bottom, transforms applied are identity, sharpening, motion blur, and two different levels of isotropic Gaussian blur. Left, middle and right columns indicate input, blur model output, and BPN output.}
\label{fig:bpn_study}
\end{figure}

We study BPN's ability to predict the parameters of the blur model (Eqn \ref{eq:blurmodel}) from image pairs. An independent test set of text ROIs from 995 frames covering 5 different scenes from both synthetic and real datasets are used. Model parameters $\psi$ are chosen at random in the following ranges: $w \in (-1, 1)$, $\sigma_x$, $\sigma_y \in (0, 5)$, $\rho \in (0,180)$, and the blur model is applied with these parameters to the ROIs to produce transformed outputs. The original and transformed images are sent through BPN and predicted parameters are compared to ground truth. Regression performance is shown in Table \ref{tab:BPN}. From observing real videos, we noted translational alignment errors in the order of 0-5 pixels between reference and non-reference ROI. Hence we report BPN performance for both ideal alignment and random simulated x-y jitter between 0-5 pixels. We note that network predictions are robust to such misalignment. The high errors for $\rho$ are presumably due to the fact that blur angle estimation becomes ill-conditioned for near-isotropic kernels. Qualitative results are shown in Fig. \ref{fig:bpn_study} confirming that BPN effectively models a variety of differential blurring and sharpening transformations between frames. 
	\begin{table}
		\centering
		\setlength{\tabcolsep}{5pt}
		\begin{tabular}{|l|c|c|c|c|}
		\hline
			 x-y misalignment & w & $\sigma_x$ & $\sigma_y$& $\rho$ \\ \hline
			 0 & 0.14 & 1.18 & 1.19 & 44.76 \\
			\hline
			+/- 5 & 0.15 & 1.18 & 1.21 & 45.3 \\
			\hline
		\end{tabular}\\[2ex]
	\caption{Mean absolute error in blur parameter predictions without and with alignment error ($\rho$ is specified in degrees).} 
	\label{tab:BPN}
	\end{table}

\subsection{Qualitative Evaluation}
\begin{figure}[t]
\centering
\includegraphics[width=6cm,height=3cm]{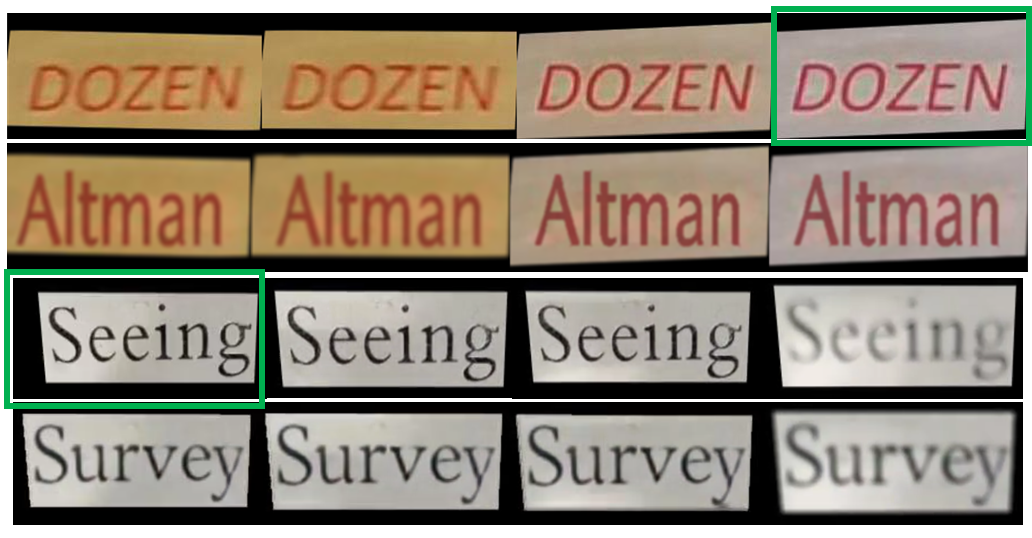}
   \caption{Results of TPM on $Robotext$ (upper half) and $Realworld$ (lower half). First/second row are original/altered video. Frames are in temporal sequence but not consecutive. Reference frame marked in green.}
\label{fig:bpn_robotic}
\end{figure}

\begin{figure}[t]
\centering
\includegraphics[width=8cm,height=5cm]{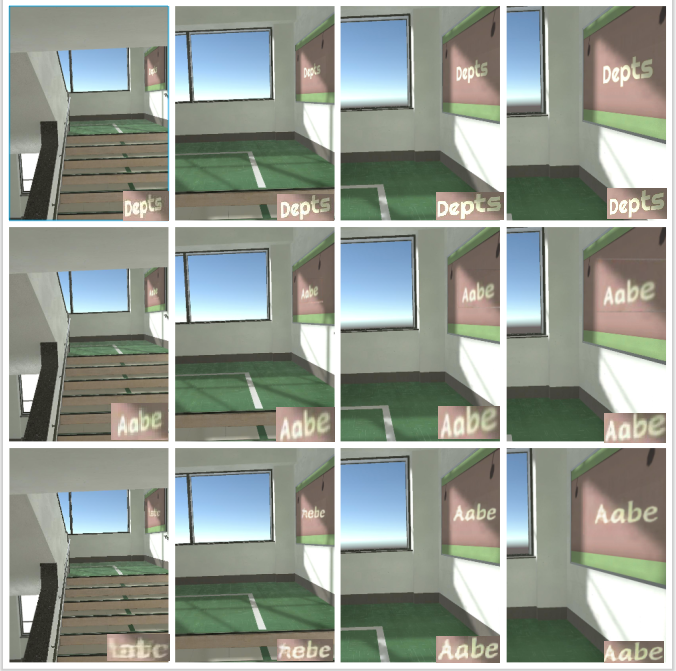}
   \caption{Comparison of original video frames (upper), STRIVE (middle) and SRNet outputs (lower) \cite{srnet2019} for a $Synthtext$ scene. Zoomed ROI insets are in lower right.}
\label{fig:comparisons_synth}
\end{figure}

\begin{figure}[t]
\centering
\includegraphics[width=8cm,height=14cm]{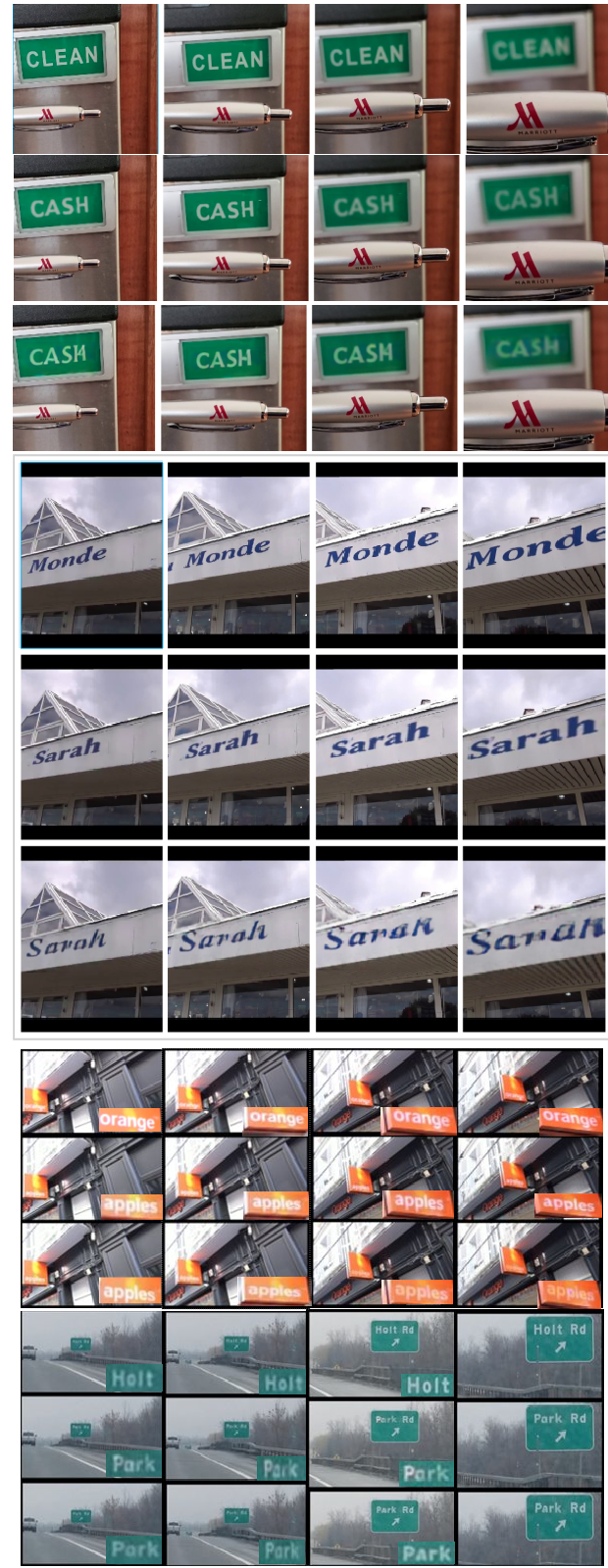}
   \caption{Comparison of original (upper), STRIVE (middle) and SRNet (lower) \cite{srnet2019} frames for real videos. Zoomed ROI insets are shown in lower right for selected scenes.}
\label{fig:comparisons_real}
\end{figure}

Fig. \ref{fig:bpn_robotic} demonstrates TPM on $Robotext$ and $Realworld$ videos. Effects of varying lighting and text sharpness as the robot moves across the scene are effectively incorporated into the altered video. Additionally we estimated blur of the text ROIs using variance of Laplacian \cite{903548}. The Pearson correlation coefficient between blur scores of original vs. replaced ROIs is 0.9912 indicating effective blur transfer. Figure \ref{fig:comparisons_synth} compares STRIVE and SRNet outputs for a $Synthtext$ video. As seen in the zoomed insets, SRNet is unable to cope with strong perspective (although being trained with such examples), and even when replacement is successful, the geometry is incorrect. STRIVE achieves both accurate text replacement and geometry by solving each task individually. Figure \ref{fig:comparisons_real} shows results for real-world videos. The first scene encounters out-of-focus blur. STRIVE is able to model this effect and maintain character integrity, while SRNet outputs are distorted. The next two scenes are from the ICDAR dataset, where videos are captured with human walking motion. The fourth scene is captured in a moving vehicle with atmospheric noise. In all cases, STRIVE is visibly superior to SRNet in preserving text integrity and the geometric and photometric effects of the original scene.
The supplemental section contains videos of challenging realistic scenarios with moving shadows, focus blur, human and vehicle motion. As seen in the videos, SRNet exhibits considerable jitter that is avoided by STRIVE, thanks to temporal smoothness constraints.


%
%


\section{Conclusions}
We propose an effective and efficient method to replace text in videos by decoupling still-image text transfer from temporal changes in geometry and appearance. The latter are modeled via a novel learnable transform that captures photometric differences between pairs of video frames. This type of differential transform learning has broad applicability for image and video editing. We offer new datasets for further progress in text-related video tasks. The efficacy of STRIVE relies on certain assumptions, including planar geometry and spectral coherence between text and background. We argue that these assumptions are not excessively limiting as such cases are commonly encountered in real world scenes. Future work includes generalizing the approach to handle occlusions, non-planar surfaces, self-lit text, and to non-text objects such as graphics and 3D shapes.

{\small
\bibliographystyle{ieee_fullname}
\bibliography{egbib}
}
\onecolumn
\begin{center}
\textbf{STRIVE: Scene Text Replacement In Videos \\
Supplemental Section}
\end{center}

\
\section{Video Demonstrations}
We include a set of 13 videos comparing the performance of STRIVE with frame-wise SRNet on scenes from our three datasets: \textit{Synthtext, Robotext, RealWorld} as summarized in Table \ref{tab:demos}. The videos can be viewed at \url{https://striveiccv2021.github.io/STRIVE-ICCV2021/}. 

\begin{table}[!htbp]
\centering
\setlength{\tabcolsep}{5pt}
\hrule
\begin{tabular}{|p{2cm}|p{2cm}|p{3.5cm}|p{8.5cm}|}
 \textbf{Filename} & \textbf{Dataset} & \textbf{Source-to-Target Text} & \textbf{Key Observations}  \\ \hline
 video1 & Synthtext & Depts to Envoy & Demonstrates importance of disentangling geometry from style. SRNet fails to replace text with strong perspective distortion.  \\
 \hline
 video2 & Synthtext & Cincinnati to Wisconsin & Demonstrates importance of temporal consistency constraints in STRIVE that produce temporally smooth output. SRNET output exhibits jitter, although the style transfer is of high quality. \\
 \hline
 video3 & Robotext & SECT to LOSE &  STRIVE preserves pose and is more temporally stable than SRNet. \\
 \hline
 video4 & RealWorld (ICDAR) & Monde to World & Performance in presence of walking motion. STRIVE exhibits superior temporal stability compared with SRNet. \\
 \hline
 video5 & RealWorld (ICDAR) & Oranges to Apples & Performance in presence of walking motion. STRIVE is temporally smooth, while SRNET output exhibits jitter, deforms the character "a" near the end, and exhibits undesirable color shifts at the text ROI boundary. \\
 \hline
 video6 & RealWorld & Ball to Prom & STRIVE is robust to complex lighting changes. Note SRNet failures towards end of clip. \\
 \hline
 video7 & RealWorld & CLEAN to CASH & STRIVE effectively predicts text blur via BPN and mimics changes in text sharpness due to depth defocus. SRNet is unable to successfully transfer text distorted by blur. \\
 \hline
 video8 & RealWorld & Seeing to Survey & STRIVE and SRNet both effectively track changes in text appearance due to depth defocus. \\
 \hline
 video9 & RealWorld & Holt to Park & Challenging noisy video with vehicle motion and atmospheric distortion. STRIVE maintains temporal consistency and character integrity, while SRNet deforms the character "a" from frame to frame. \\
 \hline
 video10 & RealWorld & COFFEE to ROBERT & STRIVE preserves geometry, while SRNet fails under strong perspective in the final segment. \\
 \hline
 video11 & RealWorld & Palm to Cold & Challenging due to small text and vehicle motion. STRIVE output is considerably smoother than SRNet. \\
 \hline
 video12 & RealWorld & OSCAR to VIOLA & STRIVE is robust to rapid nonlinear camera motion. \\
 \hline
 video13 & RealWorld & THURLBY to FAIRPOT & SRNet exhibits considerably more jitter than STRIVE \\
 \hline
\end{tabular}\\[2ex]
\caption{\textbf{Explanation of Video Demonstrations}} 
\label{tab:demos}
\end{table}

The videos cover challenging indoor and outdoor scenes with varied text styles and geometries, different types of camera and object motion, and lighting effects. Each video is a montage with the \textbf{original clip on the left, STRIVE output in the middle and SRNet framewise output on the right}. \\\\
\section{Comparisons with Pix2Pix}
\begin{figure}[htb]
\centering
\includegraphics[width=6cm]{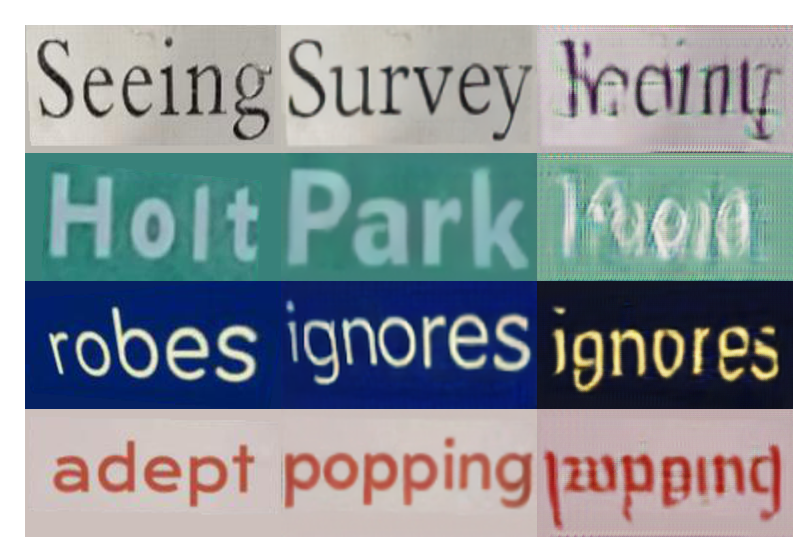}
   \caption{From left to right: Input reference frame, STRIVE, Pix2Pix. Top two are real videos, and bottom two are synthetic videos. }
\label{fig:pix2pix}
\end{figure}

In the main paper we compare STRIVE, SRNet and pix2pix on quantitative metrics in Table 1. Here we include qualitative results. Fig.  \ref{fig:pix2pix} compares STRIVE and pix2pix outputs on reference frames for 4 videos. Note that on the reference frame, STRIVE and SRNet produce identical outputs, hence are not duplicated in this figure. As seen in these examples, pix2pix does not perform well. Our explanation is that pix2pix is a general-purpose style transfer method. Although trained on text ROIs from our video datasets with temporal consistency loss over multiple frames, its architecture is not specialized for text replacement. Therefore it is unable to compete with techniques like SRNet which is explicitly designed to transfer foreground text style to new content while preserving the background. Given that pix2pix is a weak baseline, we do not include it in the video demonstrations. 

\end{document}